\documentclass[runningheads]{llncs}

\let\llncssubparagraph\subparagraph
\let\subparagraph\paragraph

\usepackage[compact]{titlesec}
\titlespacing{\section}{0pt}{2ex}{1ex}
\titlespacing{\subsection}{0pt}{1ex}{0ex}
\titlespacing{\subsubsection}{0pt}{0.5ex}{0ex}

\let\subparagraph\llncssubparagraph

\usepackage{graphicx}

\usepackage{tikz}
\usepackage{comment}
\usepackage{amsmath,amssymb} 
\usepackage{color}

\usepackage[accsupp]{accessibility}  
\usepackage{caption}
\usepackage{subcaption}

\usepackage{url}
\usepackage{hyperref}
\usepackage{numprint}
\npthousandsep{,}

\newcommand{\img}[2]{\includegraphics[width=#2\textwidth]{images/#1}}

\DeclareMathOperator*{\argmax}{arg\,max}

\begin{document}
\pagestyle{headings}
\mainmatter
\def\ECCVSubNumber{15}  

\title{Motion-Based Weak Supervision for Video Parsing with Application to Colonoscopy} 

\titlerunning{Motion-Based Weak Supervision for Video Parsing}
%
\author{Ori Kelner \and
Or Weinstein \and
Ehud Rivlin \and 
Roman Goldenberg 
}
\authorrunning{O. Kelner et al.}
%
\institute{Verily Life Sciences}
\maketitle
\vspace{-4mm}

\begin{abstract}
We propose a two-stage unsupervised approach for parsing videos into phases. We use motion cues to divide the video into coarse segments. Noisy segment labels are then used to weakly supervise an appearance-based classifier. We show the effectiveness of the method for phase detection in colonoscopy videos.

\keywords{Video, Parsing, Weak Supervision, Motion, Medical}
\end{abstract}

\section{Introduction}
We consider the problem of parsing a video into segments that belong to a small number of predefined categories. In medical imaging, e.g. surgical or other endoscopic videos, a medical procedure typically follows a sequence of known steps. Dividing a medical video into phases~\cite{garrow2021machine}
is an enabler for a number of important downstream tasks.

Here, we deal with videos collected during colonoscopy procedures. Screening colonoscopy is the gold-standard technique for removing precancerous polyps from the colon to prevent colorectal cancer. Colonoscopy is a two-phase procedure. In the first phase ("intubation") the endoscope is inserted 
all the way to the end the colon. The second phase ("withdrawal") is diagnostic, when the endoscope is slowly extracted while examining the colon to detect lesions. 

The reported colonoscopy phase detection methods~\cite{withdrawal,9680199} use supervised models to recognize the phase from a single frame and/or a local motion and then integrate the phase signal over time. The problem is that annotating videos at a frame level is extremely labor intensive, especially if the local signal is weak, as in our case. Moreover, even finding phase boundaries is not easy, as it requires medical expertise. Hence, we would like to avoid any human supervision. 

\section{Methods}

\begin{figure}[t]
    \centering
    \includegraphics[height=3.5cm, width=12.1cm]{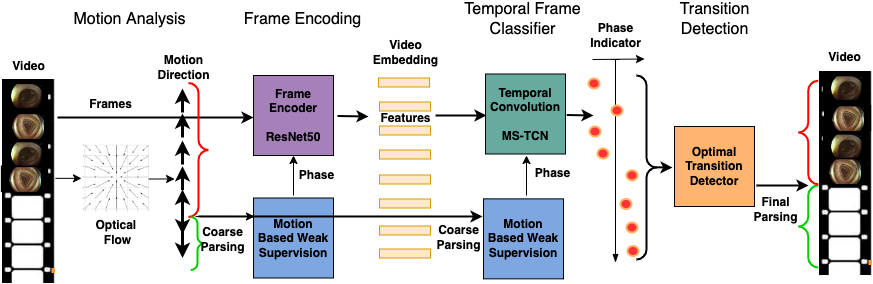} 
    \caption{Training and inference pipeline}
    \label{fig:pipeline}
\end{figure}

Our key observation is that a non-ML-based motion analysis can provide a rough estimation of phase boundaries, which can be used to weakly supervise a frame appearance classifier. The proposed training and inference pipeline is depicted in Figure~\ref{fig:pipeline}. The noisy motion-based labels are used to train a single-frame phase classifier. We encode video frames by the output of the penultimate layer of the classifier. Frame embeddings are then fed into a temporal convolution phase recognition model, also trained using the motion-based supervision. This is followed by a transition detector that finds the optimal phase boundary. We provide detailed descriptions of the pipeline components below.

\subsection{Motion-Based Parsing}

\begin{figure*}
     \centering
     \begin{subfigure}[b]{0.18\textwidth}
         \centering
         \includegraphics[width=\textwidth]{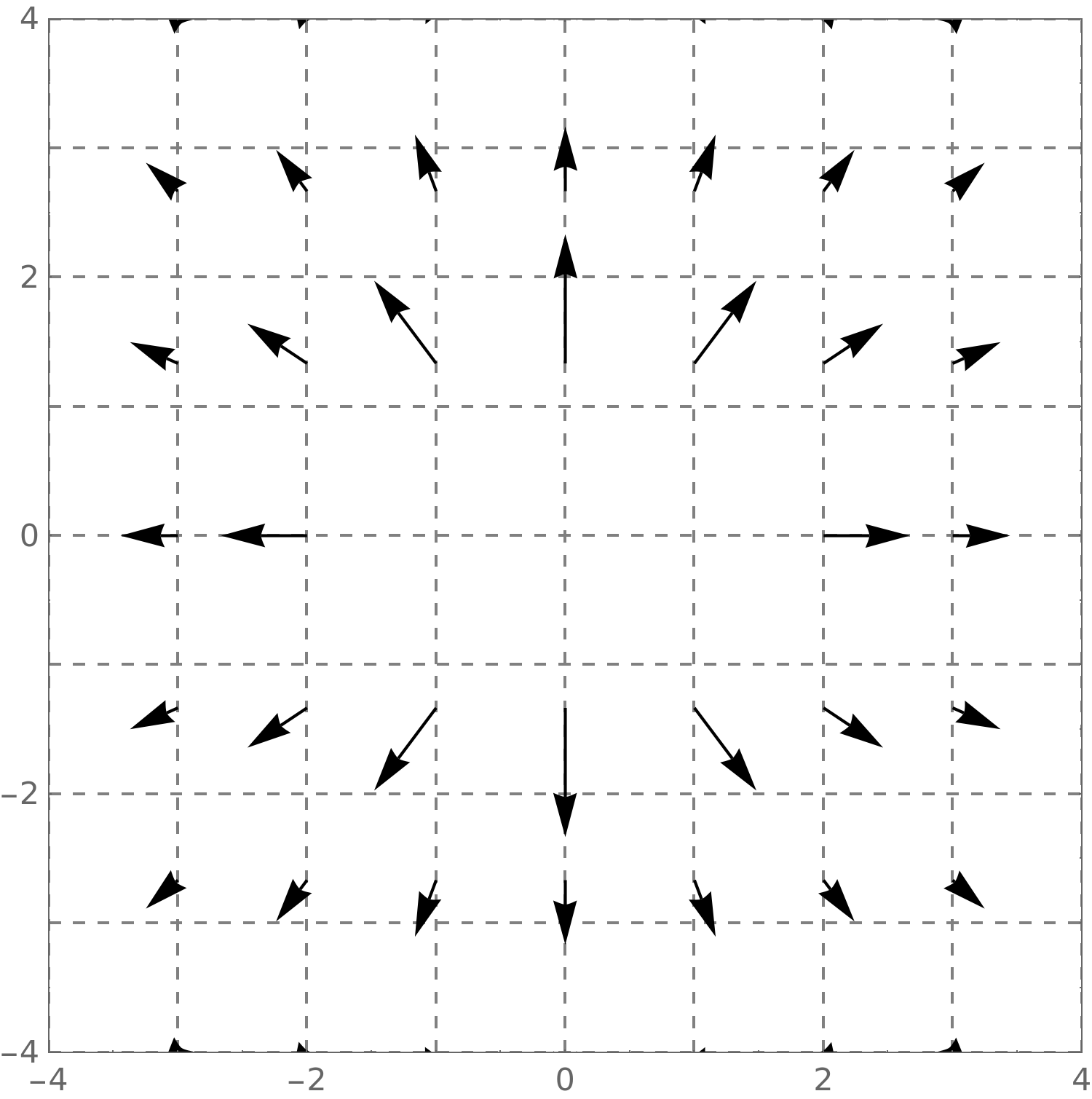}
         \caption{}
         \label{fig:y equals x}
     \end{subfigure}
     \begin{subfigure}[b]{0.18\textwidth}
         \centering
         \includegraphics[width=\textwidth]{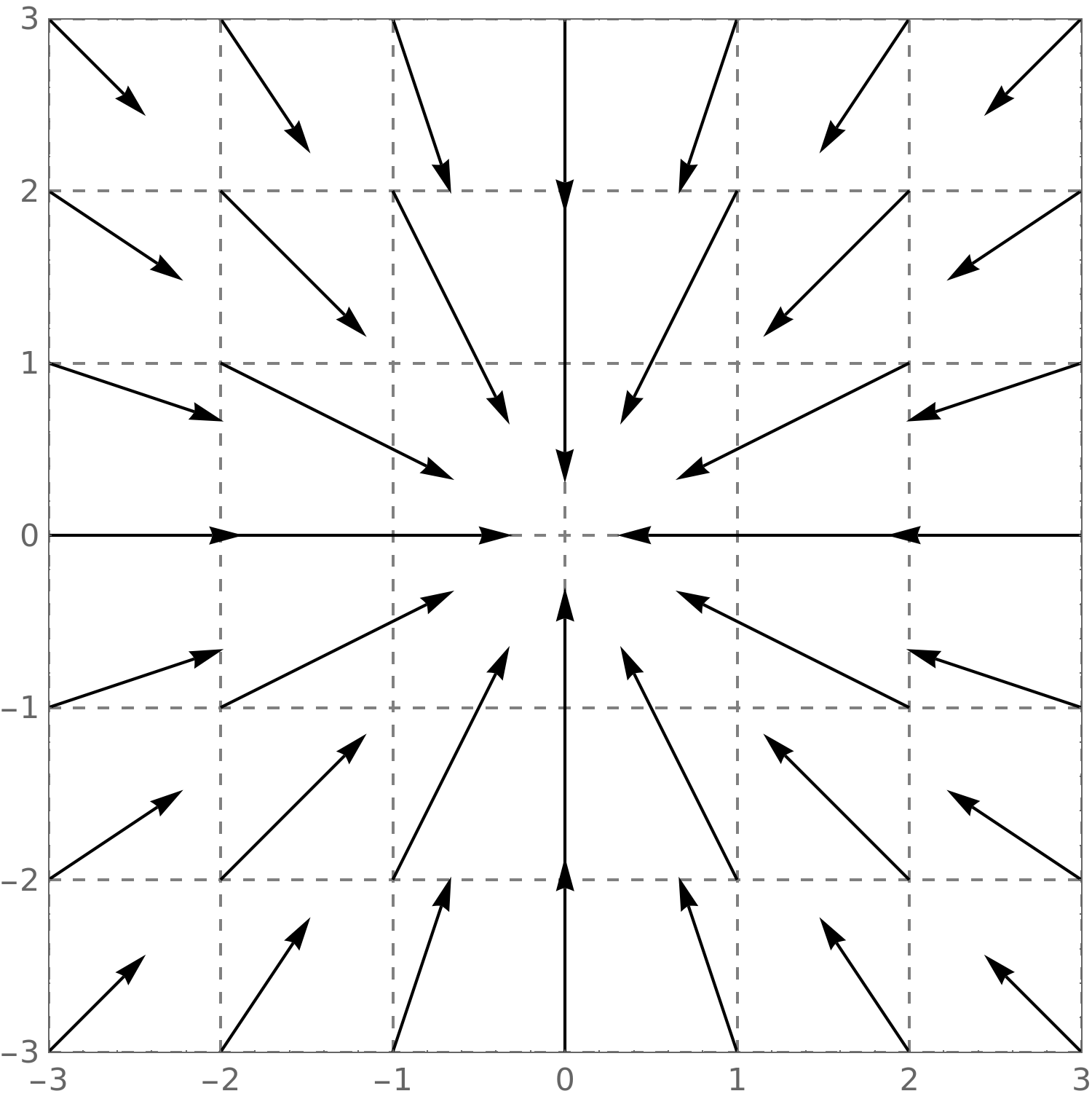}
         \caption{}
         \label{fig:three sin x}
     \end{subfigure}
     \begin{subfigure}[b]{0.18\textwidth}
         \centering
         \includegraphics[width=\textwidth]{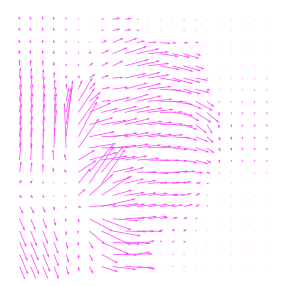}
         \caption{}
         \label{fig:five over x}
     \end{subfigure}
     \begin{subfigure}[b]{0.3\textwidth}
         \centering
         \includegraphics[width=\textwidth]{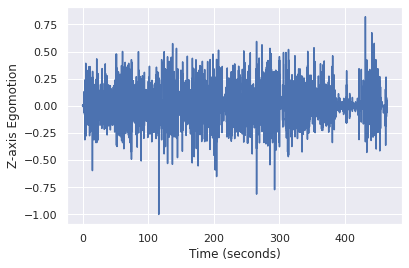}
         \caption{}
         \label{fig:six over x}
     \end{subfigure}
     \caption{(a,b) Ideal forward/backward and (c) real colonoscopy optical flow. (d) Motion direction measure for a colonoscopy video.}
     \label{fig:vector_field}
\end{figure*}

The predominant camera motion is into the colon for the intubation and out for the withdrawal. Assuming the camera is roughly aligned with the colon lumen, moving forwards indicates the intubation, and backwards - withdrawal. We use the optical flow $\mathbf{F}(x,y)$ to estimate the direction of ego-motion. Expansion/contraction of $\mathbf{F}$ indicates a forward/backward motion in the direction of the focus of expansion (Fig~\ref{fig:vector_field}a,b). The divergence $\mathbf{\nabla} \cdot \mathbf{F}=\frac{\partial F_{x}}{\partial x} + \frac{\partial F_{y}}{\partial y}$ is a differential measure of the local expansion. An integral of $\mathbf{\nabla} \cdot \mathbf{F}$ over a closed image patch $D$ is a degree of expansion measure invariant of the focus position. Following~\cite{10.5555/888701} we use Green theorem to replace the ill-conditioned divergence by a numerically stable line integral over $D$'s boundary:
$\iint_D \mathbf{\nabla} \cdot \mathbf{F} \,dx\,dy =  \oint_{\partial D} \mathbf{F} \cdot \mathbf{n} \,ds$, where $\mathbf{n}$ is the normal to the boundary $\partial D$.

\begin{figure}
\begin{subfigure}{.5\linewidth}
\centering
\img{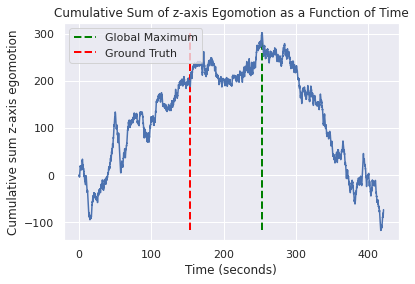}{0.6}
\label{fig:sub5}
\end{subfigure}%
\begin{subfigure}{.5\linewidth}
\centering
\img{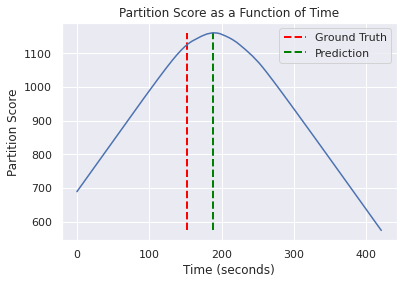}{0.6}
\label{fig:sub6}
\end{subfigure}
\caption{Cumulative motion direction vs. partition score $L(t)$ for the same video. }
\label{fig:compare}
\end{figure}

Practically, the computed $\mathbf{F}$ is very far from the radial pattern (Fig. \ref{fig:vector_field}c), which results in very noisy motion direction measurements (Fig. \ref{fig:vector_field}d).
Yet, integrating the motion direction over time, allows estimating the video phase boundary as the global maximum of the cumulative "distance" graph (Fig. \ref{fig:compare}left).

\subsection{Weakly Supervised Frame Appearance Classification}
The two colonoscopy phases differ in the nature of the performed activities. In intubation the camera is often stuck in the colon wall, whereas in withdrawal it  more often sees the colon lumen. In withdrawal the colon is inflated and polyp inspection, management, and cleansing are more frequent. We would like to leverage the distinctive frame appearance for phase detection.
\begin{figure}
    \begin{center}
        \begin{tabular}{cccc}
            \img{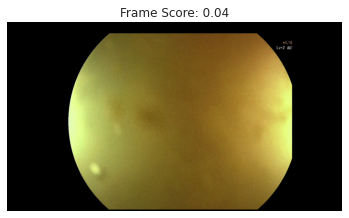}{0.3} &  
            \img{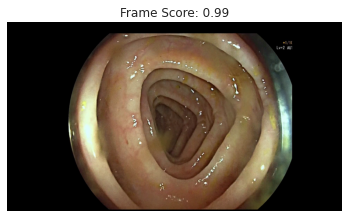}{0.3}
        \end{tabular}
    \end{center}
    \vspace{-15pt}
    \caption{Intubation and withdrawal frames and their phase prediction scores. Note that in principal it's impossible to determine the phase from a single frame as it could belong to each one of the phases.}
    \label{fig:per_frame_model}
\end{figure}

We follow a common two stage paradigm for video parsing: per frame classifier followed by temporal aggregation on features extracted by the classifier. For the second stage we use the Multi-Stage Temporal Convolutional Network (MS-TCN)~\cite{farha2019ms}. Both networks are trained using the noisy phase labels derived from the motion-based parsing (Fig. \ref{fig:pipeline}). Fig.~\ref{fig:per_frame_model} shows typical frames from the two phases and their phase prediction scores. 

\subsection{Transition Detector}
Let $\widehat{p}_{t,c}$ denote the probability estimate for phase $c$ at time $t$ generated by MS-TCN. Assuming the colonoscopy procedure is a fixed order sequence of two continuous phases, we are looking for the optimal transition point $\hat{t}$.
Similar to~\cite{withdrawal}, we define a video partition score function
$L(t)  = \sum_{\tau\in[0,t)} \widehat{p}_{\tau, 1} + \sum_{\tau\in[t,T]}\widehat{p}_{\tau, 2}$.
The optimal transition point is then $\hat{t} = \argmax_{t}{L(t)}$.

\section{Experiments}
Our training set consists of 11502 colonoscopy videos, recorded in 7 medical centers.
The evaluation set has 465 colonoscopy videos, annotated by an experienced gastroenterologist, who marked the withdrawal phase start time.
We used the Dense Inverse Search (DIS)~\cite{https://doi.org/10.48550/arxiv.1603.03590} to compute the optical flow. The frame encoder is ResNet50~\cite{DBLP:journals/corr/HeZRS15}, pre-trained on ImageNet~\cite{5206848} and fine-tuned as a frame phase classifier. For the fine-tuning, 1.5 mln frames were randomly, balanced sampled from the two phases as predicted by the motion analysis. (See Fig. \ref{fig:per_frame_model}).
The MS-TCN model was trained with 4 stages, each with 10 layers.

We compare the predicted phase transition time with the annotated, and report Mean (MAE) and Median (MedAE) Absolute Error.
We preform an ablation study to better understand the role of different components. We take the optical flow-based motion analysis as the baseline. Our proposed method is significantly better than the baseline - MAE/MedAE reduced by 47{\%} and over 50{\%} respectively. We then test the performance of the MS-TCN on different input features, including Motion Direction, ImageNet pre-trained features, and combined Classifier feature and Motion Detector (see Table~\ref{tab:ablation_metrics}). One can see that the proposed method outperforms all other options. Interestingly, adding the motion signal to MS-TCN also degrades the accuracy. A possible explanation is that the noisy frame labels are generated by the motion direction signal, and adding them to the visual features hurts the generalization capabilities of the model as it can predict the noisy labels with the motion signal itself.

\vspace{-6mm}

\begin{table}[ht]
    \small
    \centering
    \caption{Evaluation and ablation study results (in min.)}
    \begin{tabular}{|c|c|c|}
        \hline
        Method    & MAE & MedAE \\
        \hline
        Baseline - optical flow-based motion analysis & 2.15 & 0.99\\
        MS-TCN on Motion Direction & 2.6 & 1.31\\
        MS-TCN on ImageNet features & 1.36 & 0.7\\
        MS-TCN on Classifier features (\textbf{our method})  &  \textbf{1.14} & \textbf{0.49}\\    
        MS-TCN on combined Classifier and Motion Direction  & 1.61 & 0.86 \\
        \hline
    \end{tabular}

    \label{tab:ablation_metrics}
    \vspace{-5mm}
\end{table}

\section{Conclusions} 
We show how a coarse motion-based video segmentation can provide weak supervision for appearance-based  parsing models. We demonstrate the effectiveness of the method for phase detection in colonoscopy. The approach can be applied in other domains for training video parsing models without annotation.

%
%
\bibliographystyle{splncs04}
\bibliography{references}
\end{document}